%% file: aaai24.tex
\documentclass[letterpaper]{article} 
\usepackage{aaai24}  
\usepackage{times}  
\usepackage{helvet}  
\usepackage{courier}  
\usepackage[hyphens]{url}  
\usepackage{graphicx} 
\usepackage{natbib}  
\usepackage{caption} 
\usepackage{algorithm}
\usepackage{algorithmic}

\usepackage{amsmath}
\usepackage{amssymb}
\usepackage{bm}
\usepackage{array}

\usepackage{booktabs}       
\usepackage{amsfonts}       
\usepackage{nicefrac}       
\usepackage{microtype}      
\usepackage{dsfont}
\usepackage{booktabs,multirow}
\usepackage{graphics}
\usepackage{tabularx}
\nocopyright
\usepackage{newfloat}
\usepackage{listings}
\DeclareCaptionStyle{ruled}{labelfont=normalfont,labelsep=colon,strut=off} 
\floatstyle{ruled}
\newfloat{listing}{tb}{lst}{}
\floatname{listing}{Listing}
\title{Beyond Prototypes: Semantic Anchor Regularization \\ for Better Representation Learning}

\author{
    Yanqi Ge\textsuperscript{\rm 1}\equalcontrib,   
    Qiang Nie\textsuperscript{\rm 2}\equalcontrib,
    Ye Huang\textsuperscript{\rm 1},
    Yong Liu\textsuperscript{\rm 2}, \\
    Chengjie Wang\textsuperscript{\rm 2,3},
    Feng Zheng\textsuperscript{\rm 4}$^\dag$,
    Wen Li\textsuperscript{\rm 1},
    Lixin Duan\textsuperscript{\rm 1,5}\thanks{Corresponding author.}
}

\affiliations{
    \textsuperscript{\rm 1} Shenzhen Institute for Advanced Study, University of Electronic Science and Technology of China\\
    \textsuperscript{\rm 2} Tencent Youtu Lab \\ 
    \textsuperscript{\rm 3} Shanghai Jiao tong University\\
    \textsuperscript{\rm 4} Southern University of Science and Technology\\
    \textsuperscript{\rm 5} Sichuan Provincial People's Hospital\\
    geyanqiqi@gmail.com
    }

\usepackage{bibentry}

\begin{document}

\setcounter{secnumdepth}{2} 

\maketitle

\begin{abstract}
One of the ultimate goals of representation learning is to achieve compactness within a class and well-separability between classes. 
Many outstanding metric-based and prototype-based methods following the Expectation-Maximization paradigm, have been proposed for this objective. 
However, they inevitably introduce biases into the learning process, particularly with long-tail distributed training data.
In this paper, we reveal that the class prototype is not necessarily to be derived from training features and propose a novel perspective to use pre-defined class anchors serving as feature centroid to unidirectionally guide feature learning. 
However, the pre-defined anchors may have a large semantic distance from the pixel features, which prevents them from being directly applied.
To address this issue and generate feature centroid independent from feature learning, a simple yet effective Semantic Anchor Regularization (SAR) is proposed. 
SAR ensures the inter-class separability of semantic anchors in the semantic space by employing a classifier-aware auxiliary cross-entropy loss during training via disentanglement learning.
By pulling the learned features to these semantic anchors, several advantages can be attained: 1) the intra-class compactness and naturally inter-class separability, 2) induced bias or errors from feature learning can be avoided, and 3) robustness to the long-tailed problem.
The proposed SAR can be used in a plug-and-play manner in the existing models. Extensive experiments demonstrate that the SAR performs better than previous sophisticated prototype-based methods.  The implementation is available at \textit{https://github.com/geyanqi/SAR}.
\end{abstract}

\begin{figure}[t]
 \centering
 \includegraphics[width=0.88\linewidth]{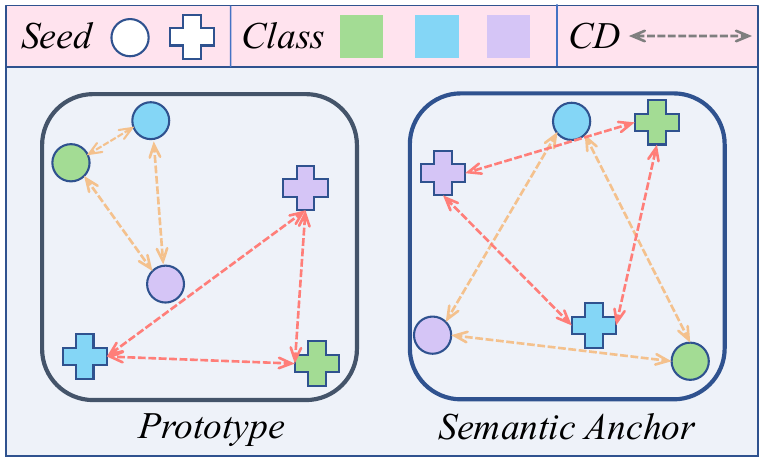}
 \caption{The difference between prototypes and semantic anchors in feature space (UMAP-Based). We train HRNet with two different seeds on Cityscapes to get these prototypes and semantic anchors. Shapes, colors, and CD represent random seeds, classes, and class dependencies, respectively.  
 The generation of semantic anchors is independent of the main task, and it achieves more consistent and weaker inter-class dependencies on imbalanced data.}
 \label{fig: seed}
\end{figure}

\section{Introduction}
\label{sec:1}
Classification, either at the image level or at the pixel level (semantic segmentation), is a foundation computer vision task with a wide range of applications, including but not limited to autonomous agent tasks such as scene understanding, augmented reality, and autonomous driving. Many efforts have been made in this problem and great progress has been achieved in recent years, especially after deep learning methods~\cite{perronnin2010improving, he2016deep, krizhevsky2017imagenet, chen2021crossvit, long2015fully, chen2017deeplab, wang2020deep} being introduced. However, no matter what kind of methods are utilized or what kind of network structures are designed, the ultimate goal is to learn representations of data that are compact within a class and separable between classes in the semantic space. To achieve this, many methods have been proposed, such as metric learning and prototype-based learning.

Metric learning is to pull together the intra-class samples and push away the samples of different categories by designing a distance metric. A lot of distance metrics have been widely utilized and benefit the representation learning, such as the contrastive loss~\cite{he2020momentum, oord2018representation,wu2018unsupervised,huang2019unsupervised,wang2020understanding,wang2021understanding,yang2022class} and the triplet loss~\cite{schroff2015facenet, ge2018deep}. 
These losses are utilized to learn effective image representations for downstream tasks by explicitly selecting positive data pairs and negative data pairs. 
~\citet{wang2020understanding} revealed that the contrastive representation benefits from the alignment of features of positive pairs and uniformity of the induced feature distribution. However, the contrastive representation relies on the construction of positive and negative sample pairs, which might induce bias in the feature learning process. 
%

Prototype-based deep learning has been attracting increasing interest recently due to its exemplar-driven nature and intuitive interpretation, which also can be deemed as only using one or several hyper-positive samples. By aligning samples with the most similar prototype in the semantic space, prototype-based methods have attained remarkable results in few-shot learning~\cite{wang2019panet,kwon2021dual}, unsupervised learning~\cite{xu2020attribute}, supervised learning~\cite{zhou2022rethinking, wang2021exploring}, and domain adaptation~\cite{jiang2022prototypical, lu2022bidirectional}, especially for long-tailed problems, \textit{e.g}, semantic segmentation. 
ProtoAttend~\cite{arik2020protoattend} shows that prototype learning is more robust when handling out-of-distribution samples, which should be attributed to the more compact data representation within the class. 
While CNN tends to learn non-discriminative features with high activations for different classes~\cite{nguyen2019feature}, i.e., the low inter-class distance. Similarly, learning more separable prototype relationships reduces the interdependence of class features, leading to enhanced generalization capabilities, especially when the training set follows a long-tailed distribution. Recently,
RegionContrast and ContrastSeg~\cite{hu2021region, wang2021exploring} propose to explore the "global" context of the training set by leveraging contrastive loss between pixel features and prototypes. CAR~\cite{huang2022car} and SASM~\cite{hong2022representation} propose directly optimizing inter-class and intra-class prototype relationships by Euclidean distance. ProtoSeg~\cite{zhou2022rethinking} proposes a non-learnable classifier using online clustering to match learned prototypes. 

However, the methods mentioned above are all via the Expectation-Maximization paradigm~\cite{moon1996expectation}, which estimates prototype assignments given learned features and updates learned features with updated prototype assignments. 
Compared to these sophisticated prototype learning methods, one realistic but seldom mentioned fact is that the relative relationships among prototypes undergo an evident drift with distinct random seeds, even though the training set and structure of the network are fixed (see Fig.~\ref{fig: seed}). 
Especially in long-tailed problems like segmentation, the prototype of the rare class appears a strong bias towards certain classes.
This phenomenon demonstrates that the traditional prototype calculations are sub-optimal since they are heavily bound to the feature learning process and distributions of training data, which can potentially result in the learning collapse for tail-end classes.

A potentially better solution could be to directly guide feature learning using well-separated and fixed class anchors.
To explore this assumption, we generate three sets of pre-defined anchors as feature centroid guiding feature learning, by randomly sampling from three distinct sources: standard normal distributions, random orthogonal matrix, and random matrix with a maximum equiangular separability structure~\cite{papyan2020prevalence}.
Subsequently, we minimize the Euclidean distance between pixel features and their corresponding anchor features to regularize the model.
Amazingly, Tab.~\ref{tab: abl} shows that although the performance of randomly generated anchors is unstable, they can be beneficial for performance sometimes, and are comparable to the performance achieved by sophisticatedly prototype-based methods. In addition, solely controlling the angular structure of these class anchors did not guarantee inter-class separability and a more noticeable performance improvement. We believe this unstable and limited improvement is due to the significant semantic gap between the randomly generated anchors and learned pixel features.

To align the anchor with features in the semantic space and keep the independence of anchor generation from feature learning, we propose a simple yet effective Semantic Anchor Regularization (SAR) for learning intra-class compact and inter-class separable representations. 
As shown in Fig.~\ref{fig: arch}, instead of collecting prototypes during feature learning process, these pre-defined class anchors $\bm{A} \in \mathbb{R}^{C \times D}$ for all categories are projected into the semantic space through a lightweight embedding layer and categorized by the classifier of the main network, where $C$ is the total class number and $D$ denotes the semantic dimension of last feature layer before classification. 
In addition, we apply two key training strategies, loss reweighting, and exponential moving average (EMA) updates, to ensure that semantic anchors obtained during training are independent of the main task. We will detail these in Sec.~\ref{sec:3}.
In addition to being supervised by GT labels, by aligning features in the main network with semantic anchors, several advantages can be achieved: 1) the intra-class compactness and inter-class separability can be intuitively achieved by pulling the feature of each class to the corresponding semantic anchor, 2) induced bias and errors of the learned prototype which is calculated as the feature center can be avoided, 3) less influenced by the number of training samples and robust in long-tailed problem. 
The main contributions of this paper are summarized as follows:
\begin{itemize}
    \item We reveal that prototype representations derived from the learned features are sub-optimal and propose a simple yet effective SAR to gain better class representation.
    \item SAR can be used in a plug-and-play manner in existing models with a little extra training cost (add $0.3$ GFLOPs and $1.56$M parameters for HRNet) and no testing cost.
    \item We evaluate the proposed approach on various challenging semantic segmentation benchmarks. Extensive experiments and visualization examples demonstrate the proposed SAR is capable of promoting intra-class compactness and inter-class separability.
\end{itemize}

\section{Related Work}
\label{sec:2}
One of the ultimate goals of learning data representation is to have good intra-class compactness and inter-class separability. In the following, we review some related works that pursue this goal in metric learning and prototype-based deep learning.

\subsection{Metric Learning} Metric learning is to pull together samples within a class and push away the samples of different categories by designing a distance metric. Among them, the contrastive loss~\cite{he2020momentum, oord2018representation,wu2018unsupervised,huang2019unsupervised,wang2020understanding,wang2021understanding}, the triplet loss~\cite{schroff2015facenet, ge2018deep}, and the $n$-pair loss~\cite{sohn2016improved} are the most widely utilized. These losses are utilized to learn effective image representations for downstream tasks by explicitly selecting positive data pairs and negative data pairs. CPC~\cite{oord2018representation} applied contrastive predictive coding to learn representations from widely different data modalities, images, speech, and natural language. MoCo~\cite{he2020momentum} proposed a momentum contrast method for unsupervised visual representation learning, which allows them to build large and consistent dictionaries. ~\citet{wang2020understanding} revealed that the contrastive representation benefits from the alignment of features from positive pairs and uniformity of the induced feature distribution. However, the contrastive representation relies on the construction of positive or negative sample pairs, which might induce bias in this process. DCL~\cite{chuang2020debiased} proposed a debiased contrastive learning method to reduce false negative samples without human annotations. After all, the ideal unbiased contrastive learning is unachievable in practice since calculating all pairwise comparisons on a large dataset is impossible.

\begin{figure*}[t]
    \!\!\centerline{\includegraphics[width=0.99\linewidth]{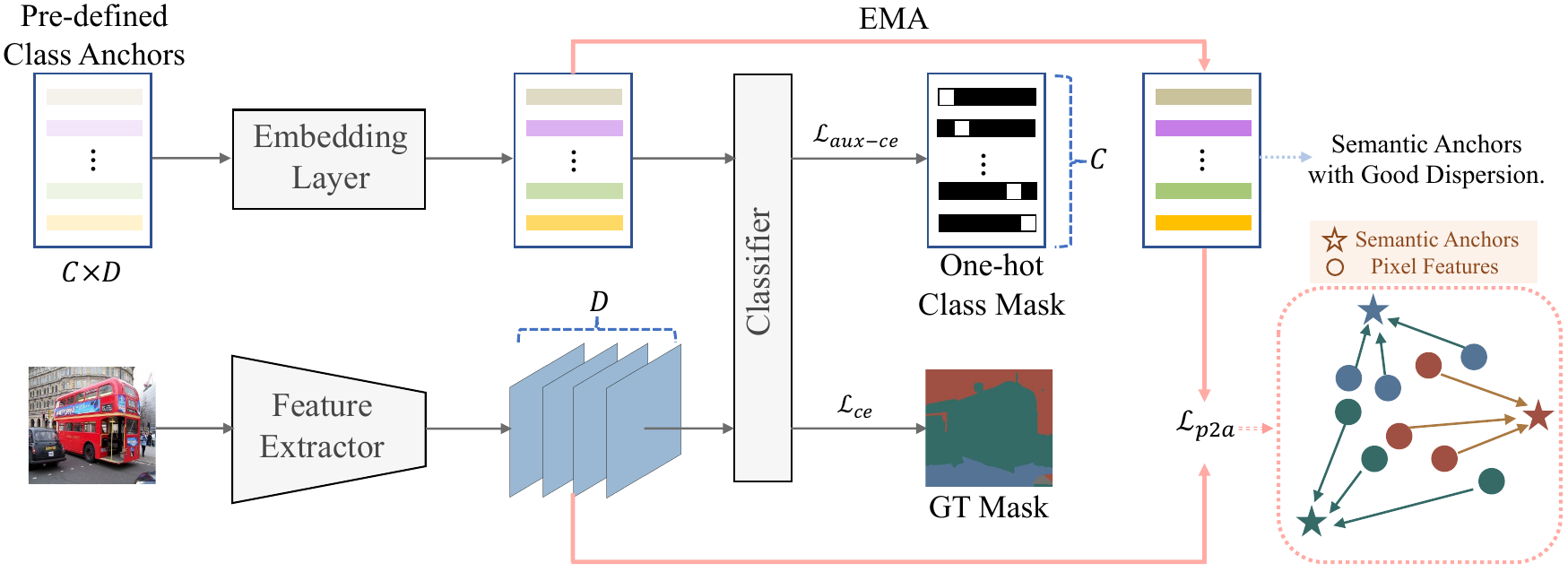}}
    \caption{Framework of the proposed method which consists of a main stream (lower stream) for segmentation/classification and an auxiliary stream (the upper stream) for SAR. Pre-defined class anchors are first embedded into the semantic space to mitigate the semantic gap and then categorized by the classifier of the mainstream. The embedded anchors are ensembled into semantic anchors in an EMA manner. The learned feature with dimension is pulled to the corresponding semantic anchor for better intra-class compactness and inter-class separability. Bold pink lines highlight the proposed SAR.}
    \label{fig: arch}
\end{figure*}

\subsection{Prototype-based Deep Learning} Prototype-based learning can be deemed as special metric learning which only considers the hyper-positive samples. Previously, prototype-based learning was combined with nearest neighbors rule~\cite{cover1967nearest} for classification tasks. Recently, a lot of work has combined prototype learning with deep neural networks and achieved remarkable results in many areas. ProtoAttend~\cite{arik2020protoattend} shows that prototype learning is more robust when handling out-of-distribution samples. DPCL~\cite{kwon2021dual} addresses the few-shot semantic segmentation problem by learning more discriminative prototypes that have larger inter-class distance and small intra-class distance in feature space. APN~\cite{xu2020attribute} utilized an attribute prototype network to transfer knowledge from known to unknown classes. To tackle the bias in calculating prototypes, BiSMAP~\cite{lu2022bidirectional} proposed multiple anisotropic prototypes. ProCA~\cite{jiang2022prototypical} proposed a prototypical contrast adaptation method for domain adaptive segmentation, which incorporates more inter-class information into class-wise prototypes. 
CAR~\cite{huang2022car} proposed optimizing representation distance from inter-class and intra-class representation relationships.
ProtoSeg~\cite{zhou2022rethinking} directly selects sub-cluster centers of embedded pixels as prototypes and implements segmentation via nonparametric nearest prototype retrieving. 
Unlike these previous methods that via EM paradigm to optimize representation relationships, SAR introduces some anchors in the semantic space to serve as feature centroids and employs them to unidirectionally guide feature learning. 
By generating feature centroids independently of feature learning, SAR is more consistent across the learning process and robust to long-tailed distribution.

\section{Method}
\label{sec:3}
\subsection{Recap of Prototype-based Deep Learning}
In the setting of semantic segmentation, each pixel $i$ of an image $I$ has to be assigned to a class $c \in C$. Specifically, let model $S_{\phi, \theta}$ comprises a feature extractor $f_{\phi}$ parameterized by $\phi$ and a classifier $g_\theta$ parameterized by $\theta$, i.e., $S_{\phi,\theta}(x) = g_\theta(f_\phi(x))$. Denote a 2D dense feature map for $I$ and its corresponding semantic feature as $\bm{F} = f_{\phi}(I) \in \mathbb{R} ^{HW \times D}$ and the ground truth label as $\bm{Y} \in \mathbb{R} ^{HW \times C}$. $H$, $W$, and $D$ denote $I$'s height and width, and number of channels, respectively. 
Existing methods typically obtain the prototype by using the average features of all pixels of a certain class during training.
Specifically, prototype $\bm{P}^c$ of a class $c$ in an image\footnote[1]{To obtain more robust prototypes, previous methods typically calculate the class centers using all the training images in a batch.} can be formulated as follows,
\begin{equation}
    \label{eq:1}
    \bm{P}^c = \frac{ \sum_{i=1}^{HW} [\bm{Y}_i == c]\!\cdot\!\bm{F} } {\sum_{i=1}^{HW} [\bm{Y}_i == c] }\in  \mathbb{R} ^{D},
\end{equation}
where $[\cdot]$ denotes the Iverson bracket.  
To improve the representation relationship between and within classes, many metric strategies $\mathcal{D}(\cdot, \cdot)$ have been proposed and can be grouped into two types: intra-class compactness loss and inter-class dispersion loss. 
The training loss with prototype regularization can be expressed as (here we take the intra-class pixel-to-prototype compactness loss as an example for illustration~\cite{huang2022car}):
\begin{equation}
    \label{eq:2}
    \mathcal{L}_{seg} = \mathcal{L}_{ce}(S_{\phi,\theta}(I), \bm{Y}) + \lambda\mathcal{D}_{intra-p2p}({\bm{Y}\!\cdot\!\bm{P}, \bm{F}}),
\end{equation}
where $\lambda$ is the trade-off that balances the cross-entropy loss $\mathcal{L}_{ce}$ and regularization loss $\mathcal{D}_{intra-c2p}$ which aims to reduce the distance between prototypes and class features. $\bm{Y}\cdot\bm{P}$ distributes  prototypes to corresponding positions in each image. Similarly, inter-class pixel-to-prototype loss can be expressed as pushing two different classes of pixel features and prototypes apart.

In the setting of classification, class prototypes can be calculated in batch data,
\begin{equation}
    \label{eq:pro_class}
    \bm{P}^c = \frac{ \sum_{i=1}^{N} [\bm{Y}_i == c]\!\cdot\!\bm{F} } { \sum_{i=1}^{N}[\bm{Y}_i == c] } \in  \mathbb{R} ^{D},
\end{equation}
where $N$ denote the batch size.

\subsection{Motivation}
\label{sec: movtivation}


Although previous prototype-based methods have achieved significant results, the following two problems still exist: 
\textit{1}) Feature entanglement. Conventionally, the prototype is generated from the learned feature and updated with consideration of the previous state (\textit{i.e.}, prototypes in memory bank)~\cite{zhou2022rethinking, wang2021exploring} during training. As a result, some errors and induced biases accumulate during the whole training process. 
For example, the bias caused by the long-tailed problem, where there are numerous features learned from red cars but very few from green cars in the training set, leads to an overemphasis on color attributes for the car's prototype.
\textit{2}) Classifier imperceptible. Although a large number of metric functions have been proposed for optimizing inter-class distance in the semantic space, they are not directly perceptible to the model classifier which predicts probabilities.

To address issue $1$, we propose Class-anchor Regularization (CR) to decouple feature centriod generation from feature learning, by pulling pixel features for each class to pre-defined class anchors with good angle relationships.
Our motivation stems from the fact that in the training paradigm of empirical risk minimization, class representations are not only bound to the data but also guided by the objective function. As seen in Fig.\ref{fig: seed}, class prototypes can be any feature vector in the semantic space as long as they are separable.
In this sense, if we explicitly guide class representations towards some pre-defined anchors that are independent of feature learning and well-separated, we can attain more consistent and discriminative class representations.
In other words, the prototype is predetermined and consistently maintains good inter-class relationships, as opposed to being estimated from the learned representations through the Expectation-Maximization (EM) paradigm. 
Errors and biases caused by long-tailed distributions can be effectively minimized compared to EM estimation.

However, as shown in Tab.~\ref{tab: abl}, CR cannot steadily improve performance since suffers from the issue $2$. The semantic gap between learned features and class anchors greatly inhibits the effect of class anchors. To solve these problems simultaneously, we further propose the classifier-aware Semantic Anchor Regularization.

\subsection{Semantic Anchor Regularization}
\label{sec: sar}
Semantic Anchor Regularization (SAR) introduces classifier-aware semantic anchors by projecting the pre-defined class anchors into the semantic space and sorting them through the classifier, to address issue $2$.
As shown in Fig.~\ref{fig: arch}, SAR learns in the fashion of multi-task learning~\cite{caruana1997multitask} by introducing a simple auxiliary steam (the upper steam) to classify the embedded anchors. The lower stream is the main task stream to perform segmentation/classification based on existing models. The $C$-way classifier is shared between the auxiliary and main streams. In training, we randomly generate pre-defined class anchors $\bm{A} \in \mathbb{R}^{C \times D}$ and fix them, and project them into the semantic space through a trainable embedding layer $h_\psi$, getting namely embedded anchors $h_\psi(\bm{A})$, and utilizing them update semantic anchors $\bm{\hat{A}}$ by Exponential Moving Average (EMA) strategy.
In this manner, the separability of semantic anchors is guaranteed according to the classifier’s decision directly in the semantic space. Hence, shifting class representations toward corresponding semantic anchors can get intra-class compact embedding space and naturally achieve inter-class separability.
Specifically, the proposed SAR is a pixel-to-anchor compactness loss by directly minimizing the distance between data representations and corresponding semantic anchors, 
\begin{equation}
    \label{eq: loss_p2a}
    \mathcal{L}_{p2a} = \mathcal{D}_{mse}(\bm{F},~ \bm{Y}\!\cdot\!\bm{\hat{A}})
\end{equation} 
The next problem that needs to be addressed is how to train embedding layers in a way that disentangles them from the main task.

\subsubsection{Disentanglement Learning.}
\label{sec: decouple}

To mitigate biased learning resulting from training drift and long-tailed distributed data, two simple yet effective training strategies are proposed to ensure that the semantic anchor is generated independently of feature learning. \textit{1}) Reweight. The classifier is required to make correct predictions with high confidence for all embedded anchors instead of the high mean confidence.
Specifically, the loss for the auxiliary task can be formulated as a weighted cross-entropy loss in Eq.~\ref{eq:aux-ce}.
\begin{equation}
    \label{eq:aux-ce}    
    \mathcal{L}_{aux-ce} = -\sum_{i=1}^C w_i \log g_\theta ^i(h_\psi(\bm{A}^i))
\end{equation}
where $w_c$ denotes the classification weight of the $c$-th pre-defined class anchor. A threshold $\tau$ is utilized to filter the high-confidence predictions in Eq.~\ref{eq: filter} and the $w_c$ can be calculated as Eq.~\ref{eq: w_c}. By re-normalizing the $w_c$ after high-confidence suppression, more attention can be put on low-confidence embedded anchors,
\begin{equation}
\label{eq: filter}
    w_c  =\begin{cases} 
        1, \quad &\text{if} \,\, g_\theta^c(h_\psi(\bm{A}^c)) > \tau \\
        {g_\theta^c(h_\psi(\bm{A}^c))}, \quad &\text{otherwise}  \\
    \end{cases}\,  
\end{equation}
\begin{equation}
\label{eq: w_c}
    w_c=\frac{\log(w_c)}{\sum_{i=1}^{C}\log(w_i)},  
\end{equation}
The above reweight strategy serves two purposes. First, it can be utilized to correct biases towards common classes the classifier learns under the guidance of the main task. Second, attributed to the Eq.~\ref{eq: filter}, embedded anchors with prediction confidence higher than $\tau$ are not changed along with the training, it can accelerate the convergence of the auxiliary task, which is already quite simple ($C$ samples, $C$-way classification), and avoid too much influence on the main task. 
In practice, for the $160$K training schedule on ADE20K~\cite{zhou2017scene}, the embedding layer is updated frequently only during the initial $600$ steps, and subsequently, it is updated approximately every $25$ steps.
\textit{2}) Update by exponential moving average. Furthermore, to avoid entangled updates of embedded anchors and main task features, we employ the Exponential Moving Average (EMA) manner to get semantic anchors at each training step $t$, 
\begin{equation}
    \label{eq:ema}
    \bm{\hat{A}}_{t} = \alpha \bm{\hat{A}}_{t-1} + (1-\alpha)h_\psi(\bm{A})_t,
\end{equation}
In addition, we only use and update semantic anchors when it is correctly classified with a probability greater than $\delta$ for better inter-class separation. 

In summary, the above training strategy ensures the independence of learning between semantic anchors and pixel features, even though the main task and auxiliary task share the same classifier, which is inherently different from previous works~\cite{huang2022car, wang2021exploring, wu2023semantic,hu2021region} collecting prototypes based on the feature learning process. 

\subsubsection{Overall.}
Integrating all components, the overall loss for SAR representation learning is the weighted sum of the presented loss components,
\begin{equation}
    \label{eq:overrall_loss}
    \mathcal{L}_{seg} = \mathcal{L}_{ce} + \lambda_1\mathcal{L}_{aux-ce} + \lambda_2\mathcal{L}_{p2a}
\end{equation}

\section{Experiments}
\label{sec: 4}
\subsection{Experimental Settings}
Semantic segmentation which is a typical and challenging classification task at the pixel level is adopted as the main downstream task to evaluate the proposed method. In addition, We further apply SAR for image classification exploratory experiment in Appendix Sec.~A.

\subsubsection{Datasets.} Our experiments are conducted on three datasets, including Cityscapes~\cite{cordts2016cityscapes}, ADE20K~\cite{zhou2017scene},  and Pascal-Context~\cite{mottaghi2014role} 
Cityscapes contains $5,\!000$ fine-grained annotated European street scenes with $2,\!975/500/1,\!524$ for train/val/test. It contains $19$ classes for scene parsing or semantic segmentation evaluation. ADE20K is one of the most challenging large-scale scene parsing datasets due to its complex scene and up to $150$ category labels. The dataset is divided into $20,\!210/2,\!000/3,\!352$ images for train/val/test, respectively. Pascal-Context is split into $4,\!998/5,\!105$ for training/test set with $59$ semantic classes plus a background class. As a common practice in semantic segmentation tasks, we use its $59$ semantic classes for evaluation. 

\subsubsection{Network Architectures.} Our implementation is based on the mmsegmentation framework~\cite{mmseg2020} and follows default model configurations. 
The embedding layer is designed as a stack of two LinearModule (Linear, Bn, ReLU) and one ConvModule (Conv, Bn, ReLU).
All backbones are initialized using corresponding weights pre-trained on ImageNet-1K~\cite{deng2009imagenet}.

\begin{table}[t]
    \centering
    \small
    \resizebox{0.78\linewidth}{!}
    {
    \begin{tabular}{cc|c}
    \toprule[1pt]
    Model & Backbone & mIoU \\
    \midrule[0.5pt]
    \midrule[0.5pt]
    FCN &  & 75.1\\
    FCN+\textbf{SAR} & \multirow{-2}{*}{ResNet-101} & \textbf{75.9} {\textbf{(+0.8)}} \\ \midrule
    DeepLabV3 & &   80.2\\
    DeepLabV3+\textbf{SAR}&  \multirow{-2}{*}{ResNet-101}  &  \textbf{80.6} {\textbf{(+0.4)}}   \\ \midrule
    HRNet &  &  79.9\\
    HRNet+\textbf{SAR} & \multirow{-2}{*}{HRNetV2-W48} &  \textbf{81.4} {\textbf{(+1.5)}} \\ \midrule
    OCRNet &  &   80.7\\
    OCRNet+\textbf{SAR}& \multirow{-2}{*}{HRNetV2-W48}&  \textbf{81.7} {\textbf{(+1.0)}}   \\ \midrule
    SegFormer &  & 81.9 \\ 
    SegFormer+\textbf{SAR} & \multirow{-2}{*}{MiT-B4} & \textbf{82.3} {\textbf{(+0.4)}} \\  \midrule
    UPerNet$^*$	& &  82.7\\
    UperNet+\textbf{SAR} & \multirow{-2}{*}{Swin-L} & \textbf{83.2} {\textbf{(+0.5)}} \\    \bottomrule[1pt]
    \end{tabular}
    }
     \caption{ Quantitative results on Cityscapes. * represents based on our reproduction.  }
    \label{tab: city}
\end{table}

\begin{table}[!hbt]
    \centering
    \small
   \resizebox{0.78\linewidth}{!}
    {
    \begin{tabular}{cc|c}
    \toprule[1pt]
    Model & Backbone & mIoU \\
    \midrule[0.5pt]
    \midrule[0.5pt]
    FCN &  & 39.9\\
    FCN+\textbf{SAR} & \multirow{-2}{*}{ResNet-101} & \textbf{40.4} {\textbf{(+0.5)}} \\ \midrule
    DeepLabV3 & &   45.0\\
    DeepLabV3+\textbf{SAR}&  \multirow{-2}{*}{ResNet-101}  &  \textbf{45.3} {\textbf{(+0.3)}}   \\ \midrule
    HRNet &  &  42.0\\
    HRNet+\textbf{SAR} & \multirow{-2}{*}{HRNetV2-W48} &  \textbf{42.8} {\textbf{(+0.8)}} \\ \midrule
    OCRNet &  &   43.2\\
    OCRNet+\textbf{SAR}& \multirow{-2}{*}{HRNetV2-W48}&  \textbf{43.7} {\textbf{(+0.5)}}   \\ \midrule
    SegFormer &  & 49.1 \\ 
    SegFormer+\textbf{SAR} & \multirow{-2}{*}{MiT-B5} & \textbf{49.5} {\textbf{(+0.4)}} \\  \midrule
    UPerNet	& & 52.2 \\
    UperNet+\textbf{SAR} & \multirow{-2}{*}{Swin-L} & \textbf{52.6} {\textbf{(+0.4)}} \\    \bottomrule[1pt]
    \end{tabular}
    }
     \caption{ Quantitative results on ADE20K.  }
    \label{tab: ade}
\end{table}

\subsubsection{Implementation Details.} The proposed SAR and its baselines use the same image augmentation for fair comparisons, including random resize with ratio $[0.5, 2.0]$, random horizontal flipping, random cropping, and random photometric distortion. 
We empirically set $\lambda_1=1$, $\alpha=0.999$, $\tau=0.9$ and $\delta=0.8$ for our all experiments. 
We use smaller $\lambda_2=0.05$ for DeepLabV3~\cite{chen2017deeplab}, which has a relatively unstable training process. In addition, to ensure generality, all other models use $\lambda_2=0.1$, although customizing hyperparameters for each benchmark can further improve performance.
Following previous work~\cite{mmseg2020, chen2017deeplab, liu2021swin}, we use the stochastic gradient descent (SGD)~\cite{robbins1951stochastic} optimizer with a learning rate of $0.01$, weight decay of $0.0005$, and momentum of $0.9$ for Convolution-based models. For Transformer-based models, we use the AdamW~\cite{loshchilov2017decoupled} optimizer with a learning rate of $0.00006$ and weight decay of $0.01$. 
The learning rate is scheduled following the polynomial annealing policy. 
For Cityscapes~\cite{cordts2016cityscapes}, we train a batch size of $8$ with a crop size of $512\times1,\!024$ (Transformer-based models trained by $1,\!024\times1,\!024$ crop size).
%
For ADE20K and Pascal-Context, we train a batch size of $16$ with a crop size of $512\times512$ and $480\times480$, respectively.
Unless otherwise specified, the models are trained for $80$k, $160$k, and $40$k iterations with 8GPUs (Transformer-based models) or 4GPUs (Convolution-based models) on Cityscapes, ADE20K, and Pascal-Context, respectively. 

\subsubsection{Evaluation Metric.} We report mean Intersection over Union (mIoU) over all classes. For fair comparisons, we do not apply any test-time data augmentation. All results reported in the baseline are derived from MMSegmentation~\cite{mmseg2020}.

\subsection{Main Results}
To verify the effectiveness, SAR is evaluated and compared with other SOTA methods on three segmentation benchmarks using different backbone networks. 

Tab.~\ref{tab: city} shows the performance on Citysacpes~\cite{cordts2016cityscapes} dataset. It can be seen that by integrating SAR with FCN~\cite{long2015fully}, DeepLabV3~\cite{chen2017rethinking}, HRNet~\cite{wang2020deep}, OCR~\cite{yuan2020object},  SegFormer~\cite{xie2021segformer} and Swin Transformer~\cite{liu2021swin}, their performance in mIoU are increased by $0.8\%$, $0.4\%$, $1.5\%$, $1.0\%$, $0.4\%$ and $0.5\%$, respectively. These improvements are significant compared to these commonly used strong baselines.

The consistent performance improvement can be observed in Tab.~\ref{tab: ade}, which adopts the same baselines as Tab.~\ref{tab: city}. In addition, we also couple SAR with DisAlign~\cite{zhang2021distribution} which is a two-stage approach specifically designed to address long-tail segmentation. We report results in Tab.~\ref{tab: ade+}, After incorporating DisAlign (DA), we achieved further improvements in the column of mTailIoU (34.5\% v.s. 34.3\%). This implies that our approach can effectively serve as a complement to methods focused on long-tailed distributions.
\begin{table}[!htb]
    \centering
    \small
    \resizebox{1\linewidth}{!}
    {
    \begin{tabular}{l|cccc}
    \toprule[1pt]
      & mIoU & mHeadIoU & mBodyIoU & mTailIoU \\ 		
    \midrule[0.5pt]
    \midrule[0.5pt]
    \multicolumn{5}{c}{\textit{Stage1}} \\
        \midrule
     HRNet	& 42.0	& 65.5	& 46.0	& 32.8 \\
     HRNet+SAR	& 42.7 (+0.7)	& 66.2 (+0.7)	& 45.6 (-0.4)	& 34.3 (+1.5) \\ 
     \midrule
     \multicolumn{5}{c}{\textit{Stage2}} \\
     \midrule
    DA+HRNet	& 42.2 (+0.2)	& 65.6 (+0.1)	& 46.0 (+0.0)	& 33.1 (+0.3)\\
    DA+SAR	& \textbf{42.9} (\textbf{+0.9})  	& 66.1 (+0.6)	& 46.0 (+0.0)& 	\textbf{34.5} (\textbf{+1.7})  \\
    \bottomrule[1pt]
    \end{tabular}
    }
    \caption{Incremental improvements for DisAlign (DA) that is focused on long-tail segmentation on ADE20K.}
    \label{tab: ade+}
\end{table}

\begin{table}[t]
    \centering
   \resizebox{0.78\linewidth}{!}
    {
    \begin{tabular}{cc|c}
    \toprule[1pt]
    Model & Backbone & mIoU (\%) \\
    \midrule[0.5pt]
    \midrule[0.5pt]
    FCN &  & 48.4\\
    FCN+\textbf{SAR} & \multirow{-2}{*}{ResNet-101} & \textbf{49.7} {\textbf{(+1.3)}} \\ \midrule
    DeepLabV3 & &   52.6\\
    DeepLabV3+\textbf{SAR}&  \multirow{-2}{*}{ResNet-101}  &  \textbf{53.3} {\textbf{(+0.7)}}   \\ \midrule
    HRNet &  &  50.3\\
    HRNet+\textbf{SAR} & \multirow{-2}{*}{HRNetV2-W48} &  \textbf{51.1} {\textbf{(+0.8)}} \\ \midrule
    OCRNet* &  &   52.0\\
    OCRNet+\textbf{SAR}& \multirow{-2}{*}{HRNetV2-W48}&  \textbf{52.4} {\textbf{(+0.4)}}   \\ \bottomrule[1pt]
    \end{tabular}
    }
     \caption{ Quantitative results on Pascal-Context. * represents based on our reproduction.  }
    \label{tab: pascal-context}
\end{table}

To show SAR's capacity for effectively handling tailed classes, we also perform experiments on Pascal-Context which follows serious long-tail distributions. The overall performance is shown in Tab.~\ref{tab: pascal-context} (MMSeg does not provide available config for Transformer-based methods on this dataset), while for a detailed analysis of specific tail-end classes, please refer to Appendix Sec.~B.

\subsection{Comparison with Prototype-based Methods}
We conduct a fair comparison between SAR and other important prototype-based methods, such as ProtoSeg~\cite{zhou2022rethinking} and CAR~\cite{huang2022car}, as these methods employ experimental settings that differ from the MMSeg benchmark. For a performance comparison of the classification task, please refer to Appendix Sec.~A.

\begin{table}[!htb]
    \centering
    \resizebox{0.750\linewidth}{!}
    {
    \begin{tabular}{ccc|c}
    \toprule[1pt]
        Method & Resolution & Schedule & mIoU  \\
    \midrule[0.5pt]
    \midrule[0.5pt]
    \multicolumn{4}{c}{\textit{Model learned on ADE20K}} \\ \midrule
    HRNet & 160K& 512$\times$512 & 42.0 \\
    SAR & 160K& 512$\times$512 & 42.8(+0.8)\\
    ProtoSeg	& 160K& 520$\times$520	& 43.0(+1.0) \\
    SAR	& 160K&  520$\times$520	& \textbf{43.3}(\textbf{+1.3}) \\
     
     \toprule[1pt]

     \multicolumn{4}{c}{\textit{Model learned on Cityscapes}} \\ 
        \midrule
        HRNet &  80K & 1024$\times$512 & 79.9 \\ 
        HRNet & 160K & 1024$\times$512 &80.6(+0.7) \\
        ProtoSeg	& 160K	& 1024$\times$512 &81.1(+1.2) \\ 
        SAR	& 80K	& 1024$\times$512 & \textbf{81.4}(\textbf{+1.5}) \\ 
     \bottomrule[1pt]
    \end{tabular}
    }
    \caption{Fair comparison of SAR and ProtoSeg based on HRNet as the baseline.}
    
    \label{tab: protoseg}
\end{table}

\begin{table}[!htb]
    \centering
    \resizebox{1\linewidth}{!}
    {
    \begin{tabular}{cc|cc|cc}
    \toprule[1pt]
    Method & mIoU & Method & mIoU &Method & mIoU\\
    \midrule[0.5pt]
    \midrule[0.5pt]
    DLV3 & 52.6 & HRNet & 50.3 & OCRNet &52.0\\ 
    CAR & 52.9 (+0.3) & CAR & 50.7 (+0.4) & CAR & 52.3 (+0.3)\\ 
    SAR & \textbf{53.3} (\textbf{+0.7}) & SAR & \textbf{51.1} (\textbf{+0.7}) & SAR & \textbf{52.5} (\textbf{+0.5}) \\
    \bottomrule[1pt]
    \end{tabular}
    }
    \caption{Fair comparison with CAR on Pascal-Context using 520$\times$520 training crops. DLV3: DeepLabV3}
    \label{tab: car}
\end{table}

\begin{figure*}[ht]    
    \centerline{\includegraphics[width=1\linewidth]{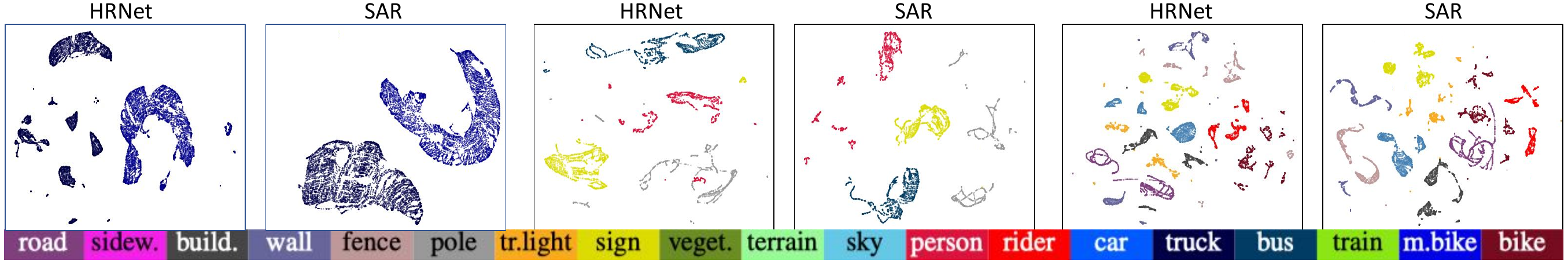}}
    %
    \caption{Visualization of the learned features with HRNet and SAR on Cityscapes utilizing UMAP.}
    \label{fig: feat_vis}
\end{figure*}

\begin{table}[!htb]
    \centering
    \small
    \resizebox{1\linewidth}{!}
    {
    \begin{tabular}{ccccc|c}
    \toprule[1pt]
     $\mathcal{L}_{ce}$ & $\mathcal{L}_{p2a}$ & $\mathcal{L}_{aux-ce}$ & EMA & Reweight  &  mIoU (\%) \\ 		
     \midrule[0.5pt]
     \midrule[0.5pt]
    $\checkmark$ & & & & & 79.9\\ 
    \midrule
     $\checkmark$& $ND$& & & & 79.8$\sim$80.3\\
     $\checkmark$& $OM$& & & & 79.2$\sim$79.9\\
     $\checkmark$& $MES$& & & & 79.8$\sim$80.4\\ 
    
    \midrule 
    $\checkmark$ & $N$  & $\checkmark$ &   &  & 80.6 ({\textbf{+0.7}})\\ 
    $\checkmark$ & $N$  & $\checkmark$ &$\checkmark$ &  & 81.1 ({\textbf{+1.2}})\\ 
    $\checkmark$ & $N$  &  $\checkmark$ & $\checkmark$& $\checkmark$& 81.4 ({\textbf{+1.5}})\\ 
    \bottomrule[1pt]
    \end{tabular}
    }
    \caption{Ablation studies on the key components of our proposed SAR on Cityscapes. $ND$: standard Normal Distribution, $OM$: random Orthogonal Matrix, $MES$: random matrix with a Maximum Equiangular Separability structure.}
    \label{tab: abl}
\end{table}

\subsection{Ablation Studies}
In Tab.~\ref{tab: abl}, we evaluate the efficacy of each component in the proposed SAR on Cityscapes~\cite{cordts2016cityscapes}. 
$\mathcal{L}_{ce}$ means the case only using HRNet as baseline. 
Without embedding layer (+$\mathcal{L}_{p2a}$), learned features in the segmentation task are directly regularized by the pre-defined anchors $A$ which are randomly sampled from the three sources.
As discussed in Sec.~\ref{sec: movtivation}, these random class anchors can improve the performance of the baseline but with strong variations. 
%
%
To reduce semantic gaps between class anchors and semantic space, we embed the pre-defined anchor into semantic space (+embedding layer) and control their separability using the classifier for segmentation (+$\mathcal{L}_{aux-ce}$). 
In this manner, a stable improvement of $0.7$ in mIoU can be obtained. Further, the EMA updating strategy and Reweighting strategy are utilized in disentanglement learning these semantic anchors. Combining all components, SAR can achieve an increment of $1.5$ in mIoU compared to the baseline.

\subsubsection{Detailed Analyses.} More detailed ablation studies can refer to Appendix Sec.~C, including independence of semantic anchor, model robustness, hyper-parameters sensitivity, and extra computational and storage burden analyses.

\subsection{Qualitative Evaluation on the Segmentation Results}
\subsubsection{Visualization of learned representations.} 
Fig.~\ref{fig: feat_vis} visualizes the feature learned with and without the proposed SAR using UMAP~\cite{mcinnes2018umap} analysis. Learning with SAR improves intra-class compactness and inter-class separability. According to the basic assumption proposed in~\cite{oliver2018realistic}, the decision boundary generated by SAR will pass through more sparse regions and have stronger robustness and generalization


\begin{figure}[!htb]
    \centerline{\includegraphics[width=1\linewidth]{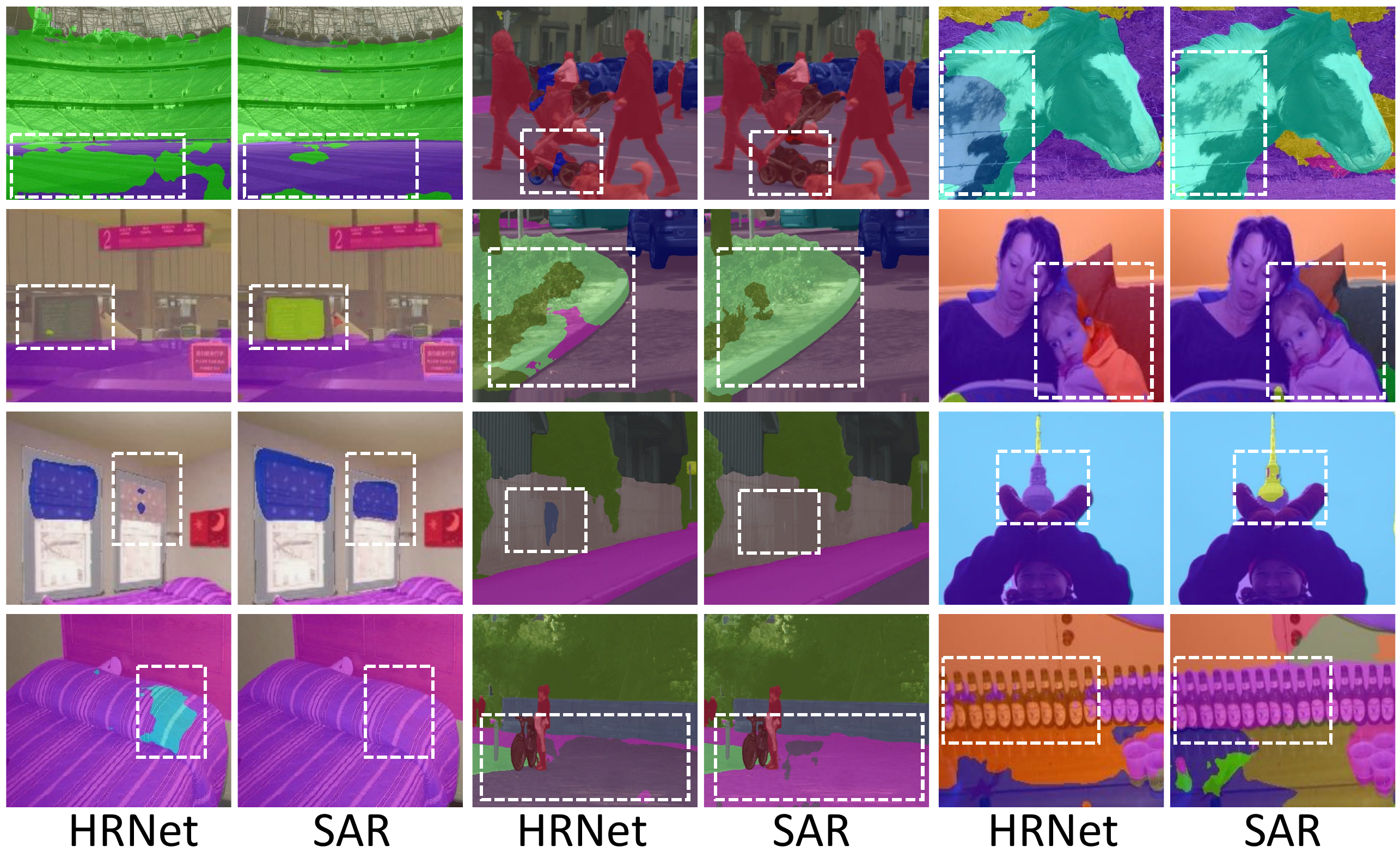}}
    \caption{Qualitative results on ADE20K (L. $2$ Cols.), Cityscapes (M. $2$ Cols.), and Pascal-Context (R. $2$ Cols.).}
    \label{fig: seg_vis}
\end{figure}

\subsubsection{Qualitative results.} 
We present qualitative examples of the segmentation results in Fig.~\ref{fig: seg_vis}. Examples are from ADE20K, Cityscapes, and Pascal-Context, respectively. The results from the HRNet and HRNet training with SAR are included for comparison. 

\section{Conclusion}
\label{sec:6}
In this paper, we present that prototype representations derived from the learned features are sub-optimal since they heavily rely on the data distribution.
We proposed a novel perspective to leverage pre-defined class anchors which are decoupled from pixel features to guide representation learning. 
However, directly using these anchors suffers from the semantic gap between pre-defined anchors and learned features in the semantic space.
To address this issue, we proposed semantic anchor regularization (SAR) for improved class representation. SAR adopts a disentangled learning approach to collect these semantic anchors, using them to unidirectionally guide feature learning.
SAR can be applied in a plug-and-play manner to help existing models achieve better performance and address long-tail distributions. 
Experiments on downstream semantic segmentation with extensive ablation studies have validated the effectiveness of the proposed SAR method. In addition, exploratory experiments in Appendix Sec.~A show SAR is promising as a general solution for classification-based tasks.
We hope that our proposal can advance future studies of representation learning and imbalanced learning. 
Limitations and future work are provided in Appendix Sec.~D. 

\section{Acknowledgments}
This work is supported by the National Natural Science Foundation of China under Grants No.~82121003 and No.~62176047, and the Shenzhen Fundamental Research Program under Grant No.~JCYJ20220530164812027.

\input{aaai24.bbl}
\clearpage

\input{appendix}

\end{document}

%% file: appendix.tex
\appendix

{\LARGE \bf \centerline{Appendix}}
\vspace{1cm}
The Appendix is organized as follows. 
Sec.~\ref{sec: cls task} further explores the application of our proposed method to the image classification task. 
Sec~\ref{sec: tailed-classes} shows the performance of SAR on rare classes.
Sec.~\ref{sec: more abl} provides detailed ablation studies of our proposed methods. Notably, in Sec.~\ref{sec: influence}, we present two clear pieces of experimental evidence supporting the semantic anchors independent of feature learning. 
Sec.~\ref{sec: discussion} discusses the limitations and future work of SAR.


\section{Application to Image Classification Task}
\label{sec: cls task}

In the exploratory experiment, we further apply SAR to the image-level classification task. We evaluate SAR on image classification performance in normal and long-tail settings.

\subsection{Normal Setting}
Our experiments are conducted on two datasets, CIFAR-100~\cite{krizhevsky2009learning} and CUB-200~\cite{wah2011caltech}. CIFAR-100 is a subset of the tiny images dataset and consists of $60,\!000$ images. The 100 classes in the CIFAR-100 are grouped into $20$ super-classes. There are 600 images per class which are split into $500$ training images and $100$ testing images per class. CUB-200 is a widely used dataset for fine-grained classification tasks. We use the CUB-200-2011 version, which contains much more pictures than the original CUB-200. It contains $11,\!788$ images of $200$ subcategories belonging to birds, $5,\!994$ for training, and $5,\!794$ for testing.

Our implementation is based on the MMPretrain framework~\cite{2023mmpretrain} and follows default model configs and training schedules. 
We use ResNet\cite{he2016deep} as the baseline model. 
Compared with the segmentation, the embedding layer is simpler and designed as two LinearModules. 
All hyper-parameters of SAR are the same as the segmentation task that we reported in the manuscript.
We use Top-1 accuracy for evaluation. The reported baseline results are derived from MMPretrain.
Tab.~\ref{table: cifar} and Tab.~\ref{table: cub} show the performance on CIFAR-100 and CUB-200 datasets, respectively. 
Through the use of SAR, our approach has shown a noticeable increase in Top-1 accuracy by $0.57$ and $0.25$ on both datasets for ResNet-50, respectively.
As for ResNet-18, we have achieved a significant improvement in Top-1 accuracy by $0.71$ and $1.06$ on both datasets, respectively.
The results demonstrate the effectiveness and potential of SAR at the image-level classification.

\subsection{Long-tailed Setting}

For long-tailed classification, we employ the CIFAR-100-LT~\cite{cao2019learning}, which is the long-tailed version of the CIFAR dataset.
It is collected by controlling the degrees of data imbalance with an imbalanced factor (IF) $\beta=\frac{N_{max}}{N_{min}}$, where $N_{max}$ and $N_{min}$ are the numbers of training samples for the most and the least frequent classes.  We conduct experiments with IF=100.

Our implementation is based on the MiSLAS~\cite{zhong2021improving} which is a two-stage approach specifically designed to address long-tailed classification, and follows default model configs (ResNet-32) and training schedules. 
Furthermore, we reproduce the CAR approach~\cite{huang2022car}, an outstanding prototype-based method, in this experiment to compare its outcomes with SAR in long-tailed classification data.
These results are summarized in Tab.~\ref{table: lt_cls}. 
By integrating SAR, MiSLAS achieves an improvement of 1.0\% mIoU over the baseline.
However, due to insufficient annotations in the classification task to support CAR in calculating reliable class centers during feature learning, along with the impact of long-tailed distributions, it becomes difficult to apply it to classification tasks.

\begin{table}[t]
    \centering
   \resizebox{0.65\linewidth}{!}
    {
    \begin{tabular}{c|c}
    \toprule[1pt]
    Method & Top-1 Acc \\
    \midrule[0.5pt]
    \midrule[0.5pt]
    ResNet-18 & 78.07\\ 
    ResNet-18+\textbf{SAR} & \textbf{78.78 } {\textbf{(+0.71)}} \\ 
    \midrule
    ResNet-50 & 79.90 \\
    ResNet-50+\textbf{SAR} &  \textbf{80.47 } {\textbf{(+0.57)}}   \\
    \bottomrule[1pt]
    \end{tabular}
    }
     \caption{ Quantitative results on CIFAR-100.  }
    \label{table: cifar}
\end{table}

\begin{table}[t]
    \centering
   \resizebox{0.65\linewidth}{!}
    {
    \begin{tabular}{c|c}
    \toprule[1pt]
    Method & Top-1 Acc \\
    \midrule[0.5pt]
    \midrule[0.5pt]
    ResNet-18 & 83.10 \\ 
    ResNet-18+\textbf{SAR} & \textbf{84.16} {\textbf{(+1.06)}}\\ 
    \midrule
    ResNet-50 & 88.19 \\
    ResNet-50+\textbf{SAR} &  \textbf{88.44} {\textbf{(+0.25)}} \\
    \bottomrule[1pt]
    \end{tabular}
    }
     \caption{ Quantitative results on CUB-200.  }
    \label{table: cub}
\end{table}

\begin{table}[!htb]
    \centering
    \resizebox{0.95\linewidth}{!}
    {
    \begin{tabular}{cc|cc}
    \toprule[1pt]
     \multicolumn{2}{c|}{Stage1} &  \multicolumn{2}{c}{Stage2} \\
     \midrule
    Method & Top-1 Acc  & Method & Top-1 Acc \\
    \midrule[0.5pt]
    \midrule[0.5pt]
    MixUp	& 39.5  &  MiSLAS	&47.0 \\ 
    MixUp+CAR	& 38.0 (-1.5) & MiSLAS	& 45.3 (-1.7)  \\ 
   MixUp+SAR	& \textbf{40.6}(\textbf{+1.1}) &  MiSLAS	& \textbf{48.0} (\textbf{+1.0}) \\
    \bottomrule[1pt]
    \end{tabular}
    }
    \caption{Quantitive results on CIFAR-100-LT. The stage 2 model is initialized by the stage 1 model of the same row in the table.}
    
    \label{table: lt_cls}
\end{table}

\subsection{Summary}
The above experiments illustrate the potential of the SAR-based perspective to become a generic component for addressing challenges posed by representation learning and long-tailed distribution problems.

\begin{table*}[t]
  \centering
\begin{minipage}[c]{0.3\linewidth}
    \centering
    \resizebox{\linewidth}{!}{\setlength{\tabcolsep}{2mm}{    
\begin{tabular}{cccc}
\toprule[1pt]
Class & HRNet & SAR & $\Delta$ \\
    \midrule[0.5pt]
    \midrule[0.5pt]
         Radiator    &  45.5  &  54.0      &    {\textbf{(+8.5)}} \\
          Glass   &  10.9   &  11.3      &  {\textbf{(+0.4)}}   \\
        Clock        &   17.3  &  23.8      &  {\textbf{(+6.5)}}   \\
         Flag     &   30.2  &    30.3    &  {\textbf{(+0.1)}}   \\
\bottomrule[1pt]
\end{tabular}
    }}
\end{minipage}
\begin{minipage}[c]{0.3\linewidth}
    \centering
    \resizebox{\linewidth}{!}{\setlength{\tabcolsep}{2mm}{
\begin{tabular}{cccc}
\toprule[1pt]
Class & HRNet & SAR & $\Delta$ \\
    \midrule[0.5pt]
    \midrule[0.5pt]
        Train  & 75.5   &   83.9   &   {\textbf{(+8.4)}} \\
        Tr.Light & 74.9     &  75.0      & {\textbf{(+0.1)}}  \\
        Rider   & 65.4   &   67.5    & {\textbf{(+2.1)}}  \\
        M.Bike & 68.2 &    68.3     &   {\textbf{(+0.1)}}  \\ 
\bottomrule[1pt]
\end{tabular}
    }}
\end{minipage}
\begin{minipage}[c]{0.29\linewidth}
    \centering
    \resizebox{\linewidth}{!}{\setlength{\tabcolsep}{2mm}{
\begin{tabular}{cccc}
\toprule[1pt]
Class & HRNet & SAR & $\Delta$ \\
    \midrule[0.5pt]
    \midrule[0.5pt]
         Cup    &   31.1   &  33.8      &  {\textbf{(+2.7)}}   \\
         Sign   &   36.3   &   39.1     &   {\textbf{(+2.8)}}  \\
         Light    &  39.7   &    40.5    &  {\textbf{(+0.8)}}  \\
        Mouse   &  34.7   &    40.3    &  {\textbf{(+5.6)}}  \\
\bottomrule[1pt]
\end{tabular}
    }}
\end{minipage}
\caption{
    Performance on the Top 4 rare classes of the ADE20K (left), Cityscapes (middle), and Pascal-Context (right), respectively.
     }
      
    \label{tab: rare classes}
\end{table*}

\section{Performance on Rare Classes}
\label{sec: tailed-classes}
Semantic segmentation is inherently a long-tailed problem. To show the SAR that decoupled from feature learning has better robustness in long-tailed problems, we report the results of Top 4 rare classes in ADE20K~\cite{zhou2017scene}, Cityscapes~\cite{cordts2016cityscapes}, and Pascal-Context~\cite{mottaghi2014role} datasets, as shown in Tab.~\ref{tab: rare classes}. SAR improves significantly in these rare classes. The above performance demonstrates the robustness of the proposed method to long-tailed problems across different datasets. As the generation of class anchors is less affected by feature learning, they are insensitive to the number of samples in different classes. In addition, the reweighting strategy in Eq.~6 ensures that the model can focus more on false predictions, which are usually tailed classes. Therefore, fewer common-case biases will be introduced from semantic anchors when serving as the feature centroid for representation learning. For example, the rarest "Mouse" class in the Pascal-Context dataset accounts for only $7\times10^{-3}$\% of the entire dataset. SAR improves the IoU of "Mouse" by 5.6\% to 40.3\%. 


\section{Detailed Ablation Studies}
\label{sec: more abl}
\subsection{Independence of Semantic Anchors.}
\label{sec: influence}

In addition to the disentanglement learning analysis mentioned in Sec.~3.3 and the SAR's capacity to address long-tail problems, there are three more straightforward experimental phenomenons that demonstrate the independence of semantic anchors from feature learning.

\subsubsection{Segmentation on extremely limited data.} 
As mentioned in Sec.~\ref{sec: tailed-classes} the “Mouse” class has an extremely rare appearance with $7\times10^{-3}$\% probability in Pascal-Context, we observe that the "Mouse" class was never predicted correctly in the training results of the three DeepLabV3 with different random seeds.
Tab.~\ref{tab: mouse1} shows the performance of SAR on the "Mouse" class with the same seed. 
This demonstrates the independence between SAR and learned features, as the learned features do not include the effective recognition features for "Mouse".

\begin{table}[!htb]
    \centering
   \resizebox{0.8\linewidth}{!}
    {
    \begin{tabular}{c|c|c}
    \toprule[1pt]
    Seed & DeepLabV3 & SAR \\
    \midrule[0.5pt]
    \midrule[0.5pt]
   1270964153 & 0.0 & \textbf{23.4 (+23.4)}\\
    1024 & 0.0 & \textbf{25.6 (+25.6)} \\
    5555 & 0.0 & \textbf{35.4 (+35.4)} \\ 
    \bottomrule[1pt]
    
    \end{tabular}
    }
     \caption{ Quantitative results (IoU) on the "Mouse" class. Our result is based on DeepLabV3 with SAR.}
    \label{tab: mouse1}
\end{table}

\begin{table}[!htb]
    \centering
   \resizebox{0.43\linewidth}{!}
    {
    \begin{tabular}{c|c}
    \toprule[1pt]
    Model & IoU \\
    \midrule[0.5pt]
    \midrule[0.5pt]
  HRNet & 34.7 \\
  CAR &  0.0 (-34.7)\\ 
  \textbf{SAR} & \textbf{40.3 (+5.6)} \\ 
    \bottomrule[1pt]
    
    \end{tabular}
    }
     \caption{ Comparing the IoU of SAR and CAR on the "Mouse" class. }
    \label{tab: mouse2}
\end{table}

\subsubsection{Compare with CAR on the "Mouse" class.}

An instance is present in Tab.~\ref{tab: mouse2}, we compared the performance of our method with CAR on the "Mouse" class. CAR is an excellent prototype-based method that calculates prototypes on learned features. However, due to the extremely low frequency of the “Mouse” class, the accumulation of error and bias causes the training of CAR to collapse in this class. On the contrary, since SAR is independent of feature learning, it actually improves the performance of the "Mouse" class.

\subsection{Robustness to Network Initialization}
Our method is correlated with the baseline model and robust to the network initialization. Tab.~\ref{tab: city_error} and Tab.~\ref{tab: pascal_error} show the performance of multiple seeds on two benchmarks, respectively. As the result shows, our method consistently improves the mIoU over its baseline using different random seeds, which demonstrates the effectiveness and robustness of SAR.

\begin{table}[!htb]
    \centering
   \resizebox{0.7\linewidth}{!}
    {
    \begin{tabular}{c|c|c}
    \toprule[1pt]
    Seed & HRNet & SAR \\
    \midrule[0.5pt]
    \midrule[0.5pt]
   1270964153 & 79.9 & \textbf{81.4 (+1.5)}\\
    1024 & 79.8 & \textbf{81.0 (+1.2)} \\
    5555 & 78.9 & \textbf{80.1 (+1.2)} \\ 
    \bottomrule[1pt]
    
    \end{tabular}
    }
     \caption{ Error analyze HRNet on Cityscapes.}
    \label{tab: city_error}
\end{table}

\begin{table}[!htb]
    \centering
   \resizebox{0.75\linewidth}{!}
    {
    \begin{tabular}{c|c|c}
    \toprule[1pt]
    Seed & DeepLabV3 & SAR \\
    \midrule[0.5pt]
    \midrule[0.5pt]
   1270964153 & 52.6 & \textbf{53.3 (+0.7)}\\
    1024 & 52.4 & \textbf{53.4 (+1.0)} \\
    5555 & 52.6 & \textbf{53.3 (+0.7)} \\ 
    \bottomrule[1pt]
    
    \end{tabular}
    }
     \caption{ Error analyze DeepLabV3 on Pascal-Context.}
    \label{tab: pascal_error}
\end{table}

\subsection{Hyper-parameter Analysis}

We conduct ablation experiments on the hyper-parameters of HRNet on Cityscapes. Tab.~\ref{tab: abl_lambda1} and Tab.~\ref{tab: abl_lambda2} summarizes the influence of hyper-parameters $\lambda_1$ and $\lambda_2$ to model performance, respectively.
It can be observed that the model performance is robust to the two trade-offs which balance the effect of the proposed auxiliary cross-entropy loss and pixel-to-anchor loss. 

\begin{table}[!htb]
    \centering
   \resizebox{0.5\linewidth}{!}
    {
    \begin{tabular}{c|ccc}
    \toprule[1pt]
    $\lambda_1$ &  0.5 & 1 & 2 \\ \midrule
    mIoU &80.9  & \textbf{81.4} & 81.3 \\
    \bottomrule[1pt]
    \end{tabular}
    }
     \caption{ Sensitivity to $\lambda_1$ on Cityscapes.}
    \label{tab: abl_lambda1}
\end{table}

\begin{table}[!htb]
    \centering
   \resizebox{0.5\linewidth}{!}
    {
    \begin{tabular}{c|ccc}
    \toprule[1pt]
    $\lambda_2$ &  0.05 & 0.1 & 0.2 \\ \midrule
    mIoU & 81.2 & \textbf{81.4} & 81.0\\  
    \bottomrule[1pt]
    \end{tabular}
    }
     \caption{ Sensitivity to $\lambda_2$ on Cityscapes.}
    \label{tab: abl_lambda2}
\end{table}

Tab.~\ref{tab: abl_confidence1} and Tab.~\ref{tab: abl_confidence2} show studies on $\tau$ for the auxiliary loss reweighting and $\delta$ for class anchors update strategies, respectively.
The $\tau$ filters class anchors with prediction confidence higher than it and makes the model put more attention on anchors that have lower confidence. However, a low $\tau$ leads the model to ignore some embedded anchors with not so high classification confidence, which means their inter-class distance to other class anchors is underoptimized.
As a result, anchors are not dispersedly distributed in the semantic space and the inter-class distance between anchors might be not well. 
With a proper $\tau$, more attention can be put on low-confidence anchors and broadening the inter-class distance between those not well-separated anchors.
The $\delta$ determines whether an embedded anchor is used as the regularization for feature learning. Similarly, for $\delta$, a high threshold ensures the learned feature is only regularized by those anchors with good inter-class separability.
A value of $\delta$ less than $\tau$ means that class anchors with confidence between $\delta$ and  $\tau$ are continuously optimized and utilized as regularization.

\begin{table}[!htb]
    \centering
   \resizebox{0.6\linewidth}{!}
    {
    \begin{tabular}{c|cccc}
    \toprule[1pt]
$\tau$ & 0.5  & 0.7 & $0.9$ & $1$ \\ \midrule
mIoU & 80.4 & 81.2 & \textbf{81.4} &81.0\\ 
    \bottomrule[1pt]
    
    \end{tabular}
    }
     \caption{ Sensitivity to $\tau$ on Cityscapes.}
    \label{tab: abl_confidence1}
\end{table}

\begin{table}[!htb]
    \centering
   \resizebox{0.6\linewidth}{!}
    {
    \begin{tabular}{c|cccc}
    \toprule[1pt]
$\delta$ & $0.5$ & $0.7$ & $0.8$ & $0.9$ \\ \midrule
mIoU & 80.3 & 81.3 & \textbf{81.4}  & 81.0\\  
    \bottomrule[1pt]
    \end{tabular}
    }
     \caption{ Sensitivity to $\delta$ on Cityscapes.}
    \label{tab: abl_confidence2}
\end{table}

\subsection{Computational and Storage Burden}
SAR requires conducting an auxiliary task during training, which brings additional training parameters. 
However, in practice, the process only imposes a minor computational and storage burden (See Tab.~\ref{tab: burden}). 
Compared to the original HRNet, our method only adds $0.03$GFLOPs and $1.56$M (2.3\%) training parameters when input images have a size of $1024 \times 1024$. 

\begin{table}[!htb]
	\centering
        \resizebox{1\linewidth}{!}{
		\begin{tabular}{c|cc|cc}
        \toprule[1pt]
        Model & Flops (GFLOPs) & $\Delta_1$ & Params (M) & $\Delta_2$ \\ \midrule
        HRNet &  374.34 & & 65.86 & \\ \midrule
        HRNet+SAR & 374.37 & 0.03 & 67.42  & 1.56 \\ 
        \bottomrule[1pt]
\hline
    \end{tabular}
}
\caption{Comparison of the Computation and storage burden on input size as $1,\!024 \times 1,\!024$}
\label{tab: burden}
\end{table}

\section{Discussion}
\label{sec: discussion}
\subsection{Limitations}
\label{sec: limitations}

In this work, we did not explore the application of SAR to object detection and query-based segmentation methods. We did not use these semantic anchors during the testing phase to ensure speed, but they may be beneficial for the performance during testing.

\subsection{Future Work}
\label{sec: future}
The key insight of this study is that the features utilized to regularize feature learning do not necessarily come from the task being trained. 
This enables us to integrate external controlled information to regularize or reinforce the training task, which is in line with the main idea of the now popular multi-modal recognition~\cite{girdhar2023imagebind, radford2021learning}.
Hence, 1) constructing semantic anchors from a multi-modal perspective to organize embedding space presumably further enhances the representation capability of the model.
2) In addition, using these semantic anchors as additional information during inference through a query-based classification idea.
3) Given the effectiveness of our approach in the classification task, initializing the segmentation model with weights pre-trained under SAR, and training the segmentation model using SAR may lead to a synergistic effect where the whole is greater than the sum of its parts. 

%% file: aaai24.bbl
\begin{thebibliography}{56}
\providecommand{\natexlab}[1]{#1}

\bibitem[{Arik and Pfister(2020)}]{arik2020protoattend}
Arik, S.~{\"O}.; and Pfister, T. 2020.
\newblock Protoattend: Attention-based prototypical learning.
\newblock \emph{The Journal of Machine Learning Research}, 21(1): 8691--8725.

\bibitem[{Cao et~al.(2019)Cao, Wei, Gaidon, Arechiga, and Ma}]{cao2019learning}
Cao, K.; Wei, C.; Gaidon, A.; Arechiga, N.; and Ma, T. 2019.
\newblock Learning imbalanced datasets with label-distribution-aware margin loss.
\newblock \emph{Advances in neural information processing systems}, 32.

\bibitem[{Caruana(1997)}]{caruana1997multitask}
Caruana, R. 1997.
\newblock Multitask learning.
\newblock \emph{Machine learning}, 28: 41--75.

\bibitem[{Chen, Fan, and Panda(2021)}]{chen2021crossvit}
Chen, C.-F.~R.; Fan, Q.; and Panda, R. 2021.
\newblock Crossvit: Cross-attention multi-scale vision transformer for image classification.
\newblock In \emph{Proceedings of the IEEE/CVF international conference on computer vision}, 357--366.

\bibitem[{Chen et~al.(2017{\natexlab{a}})Chen, Papandreou, Kokkinos, Murphy, and Yuille}]{chen2017deeplab}
Chen, L.-C.; Papandreou, G.; Kokkinos, I.; Murphy, K.; and Yuille, A.~L. 2017{\natexlab{a}}.
\newblock Deeplab: Semantic image segmentation with deep convolutional nets, atrous convolution, and fully connected crfs.
\newblock \emph{TPAMI}, 40(4): 834--848.

\bibitem[{Chen et~al.(2017{\natexlab{b}})Chen, Papandreou, Schroff, and Adam}]{chen2017rethinking}
Chen, L.-C.; Papandreou, G.; Schroff, F.; and Adam, H. 2017{\natexlab{b}}.
\newblock Rethinking atrous convolution for semantic image segmentation.
\newblock \emph{arXiv preprint arXiv:1706.05587}.

\bibitem[{Chuang et~al.(2020)Chuang, Robinson, Lin, Torralba, and Jegelka}]{chuang2020debiased}
Chuang, C.-Y.; Robinson, J.; Lin, Y.-C.; Torralba, A.; and Jegelka, S. 2020.
\newblock Debiased contrastive learning.
\newblock \emph{Advances in neural information processing systems}, 33: 8765--8775.

\bibitem[{Contributors(2020)}]{mmseg2020}
Contributors, M. 2020.
\newblock {MMSegmentation}: OpenMMLab Semantic Segmentation Toolbox and Benchmark.
\newblock \url{https://github.com/open-mmlab/mmsegmentation}.

\bibitem[{Contributors(2023)}]{2023mmpretrain}
Contributors, M. 2023.
\newblock OpenMMLab's Pre-training Toolbox and Benchmark.
\newblock \url{https://github.com/open-mmlab/mmpretrain}.

\bibitem[{Cordts et~al.(2016)Cordts, Omran, Ramos, Rehfeld, Enzweiler, Benenson, Franke, Roth, and Schiele}]{cordts2016cityscapes}
Cordts, M.; Omran, M.; Ramos, S.; Rehfeld, T.; Enzweiler, M.; Benenson, R.; Franke, U.; Roth, S.; and Schiele, B. 2016.
\newblock The cityscapes dataset for semantic urban scene understanding.
\newblock In \emph{CVPR}, 3213--3223.

\bibitem[{Cover and Hart(1967)}]{cover1967nearest}
Cover, T.; and Hart, P. 1967.
\newblock Nearest neighbor pattern classification.
\newblock \emph{IEEE transactions on information theory}, 13(1): 21--27.

\bibitem[{Deng et~al.(2009)Deng, Dong, Socher, Li, Li, and Fei-Fei}]{deng2009imagenet}
Deng, J.; Dong, W.; Socher, R.; Li, L.-J.; Li, K.; and Fei-Fei, L. 2009.
\newblock Imagenet: A large-scale hierarchical image database.
\newblock In \emph{CVPR}, 248--255.

\bibitem[{Ge(2018)}]{ge2018deep}
Ge, W. 2018.
\newblock Deep metric learning with hierarchical triplet loss.
\newblock In \emph{Proceedings of the European Conference on Computer Vision (ECCV)}, 269--285.

\bibitem[{Girdhar et~al.(2023)Girdhar, El-Nouby, Liu, Singh, Alwala, Joulin, and Misra}]{girdhar2023imagebind}
Girdhar, R.; El-Nouby, A.; Liu, Z.; Singh, M.; Alwala, K.~V.; Joulin, A.; and Misra, I. 2023.
\newblock ImageBind: One Embedding Space To Bind Them All.
\newblock In \emph{Proceedings of the IEEE/CVF Conference on Computer Vision and Pattern Recognition}, 15180--15190.

\bibitem[{He et~al.(2020)He, Fan, Wu, Xie, and Girshick}]{he2020momentum}
He, K.; Fan, H.; Wu, Y.; Xie, S.; and Girshick, R. 2020.
\newblock Momentum contrast for unsupervised visual representation learning.
\newblock In \emph{Proceedings of the IEEE/CVF conference on computer vision and pattern recognition}, 9729--9738.

\bibitem[{He et~al.(2016)He, Zhang, Ren, and Sun}]{he2016deep}
He, K.; Zhang, X.; Ren, S.; and Sun, J. 2016.
\newblock Deep residual learning for image recognition.
\newblock In \emph{CVPR}, 770--778.

\bibitem[{Hong et~al.(2022)Hong, Pan, Sun, Yu, and Gao}]{hong2022representation}
Hong, Y.; Pan, H.; Sun, W.; Yu, X.; and Gao, H. 2022.
\newblock Representation Separation for Semantic Segmentation with Vision Transformers.
\newblock \emph{arXiv preprint arXiv:2212.13764}.

\bibitem[{Hu, Cui, and Wang(2021)}]{hu2021region}
Hu, H.; Cui, J.; and Wang, L. 2021.
\newblock Region-aware contrastive learning for semantic segmentation.
\newblock In \emph{Proceedings of the IEEE/CVF International Conference on Computer Vision}, 16291--16301.

\bibitem[{Huang et~al.(2019)Huang, Dong, Gong, and Zhu}]{huang2019unsupervised}
Huang, J.; Dong, Q.; Gong, S.; and Zhu, X. 2019.
\newblock Unsupervised deep learning by neighbourhood discovery.
\newblock In \emph{International Conference on Machine Learning}, 2849--2858. PMLR.

\bibitem[{Huang et~al.(2022)Huang, Kang, Chen, Zhe, Jia, Bao, and He}]{huang2022car}
Huang, Y.; Kang, D.; Chen, L.; Zhe, X.; Jia, W.; Bao, L.; and He, X. 2022.
\newblock Car: Class-aware regularizations for semantic segmentation.
\newblock In \emph{Computer Vision--ECCV 2022: 17th European Conference, Tel Aviv, Israel, October 23--27, 2022, Proceedings, Part XXVIII}, 518--534. Springer.

\bibitem[{Jiang et~al.(2022)Jiang, Li, Yang, Gao, Wang, Tai, and Wang}]{jiang2022prototypical}
Jiang, Z.; Li, Y.; Yang, C.; Gao, P.; Wang, Y.; Tai, Y.; and Wang, C. 2022.
\newblock Prototypical contrast adaptation for domain adaptive semantic segmentation.
\newblock In \emph{Computer Vision--ECCV 2022: 17th European Conference, Tel Aviv, Israel, October 23--27, 2022, Proceedings, Part XXXIV}, 36--54. Springer.

\bibitem[{Krizhevsky, Hinton et~al.(2009)}]{krizhevsky2009learning}
Krizhevsky, A.; Hinton, G.; et~al. 2009.
\newblock Learning multiple layers of features from tiny images.

\bibitem[{Krizhevsky, Sutskever, and Hinton(2017)}]{krizhevsky2017imagenet}
Krizhevsky, A.; Sutskever, I.; and Hinton, G.~E. 2017.
\newblock Imagenet classification with deep convolutional neural networks.
\newblock \emph{Communications of the ACM}, 60(6): 84--90.

\bibitem[{Kwon et~al.(2021)Kwon, Jeong, Kim, and Sohn}]{kwon2021dual}
Kwon, H.; Jeong, S.; Kim, S.; and Sohn, K. 2021.
\newblock Dual prototypical contrastive learning for few-shot semantic segmentation.
\newblock \emph{arXiv preprint arXiv:2111.04982}.

\bibitem[{Liu et~al.(2021)Liu, Lin, Cao, Hu, Wei, Zhang, Lin, and Guo}]{liu2021swin}
Liu, Z.; Lin, Y.; Cao, Y.; Hu, H.; Wei, Y.; Zhang, Z.; Lin, S.; and Guo, B. 2021.
\newblock Swin transformer: Hierarchical vision transformer using shifted windows.
\newblock In \emph{Proceedings of the IEEE/CVF international conference on computer vision}, 10012--10022.

\bibitem[{Long, Shelhamer, and Darrell(2015)}]{long2015fully}
Long, J.; Shelhamer, E.; and Darrell, T. 2015.
\newblock Fully convolutional networks for semantic segmentation.
\newblock In \emph{Proceedings of the IEEE conference on computer vision and pattern recognition}, 3431--3440.

\bibitem[{Loshchilov and Hutter(2017)}]{loshchilov2017decoupled}
Loshchilov, I.; and Hutter, F. 2017.
\newblock Decoupled weight decay regularization.
\newblock \emph{arXiv preprint arXiv:1711.05101}.

\bibitem[{Lu et~al.(2022)Lu, Luo, Zhang, Li, Yang, and Xiao}]{lu2022bidirectional}
Lu, Y.; Luo, Y.; Zhang, L.; Li, Z.; Yang, Y.; and Xiao, J. 2022.
\newblock Bidirectional self-training with multiple anisotropic prototypes for domain adaptive semantic segmentation.
\newblock In \emph{Proceedings of the 30th ACM International Conference on Multimedia}, 1405--1415.

\bibitem[{McInnes, Healy, and Melville(2018)}]{mcinnes2018umap}
McInnes, L.; Healy, J.; and Melville, J. 2018.
\newblock Umap: Uniform manifold approximation and projection for dimension reduction.
\newblock \emph{arXiv}.

\bibitem[{Moon(1996)}]{moon1996expectation}
Moon, T.~K. 1996.
\newblock The expectation-maximization algorithm.
\newblock \emph{IEEE Signal processing magazine}, 13(6): 47--60.

\bibitem[{Mottaghi et~al.(2014)Mottaghi, Chen, Liu, Cho, Lee, Fidler, Urtasun, and Yuille}]{mottaghi2014role}
Mottaghi, R.; Chen, X.; Liu, X.; Cho, N.-G.; Lee, S.-W.; Fidler, S.; Urtasun, R.; and Yuille, A. 2014.
\newblock The role of context for object detection and semantic segmentation in the wild.
\newblock In \emph{Proceedings of the IEEE conference on computer vision and pattern recognition}, 891--898.

\bibitem[{Nguyen and Todorovic(2019)}]{nguyen2019feature}
Nguyen, K.; and Todorovic, S. 2019.
\newblock Feature weighting and boosting for few-shot segmentation.
\newblock In \emph{Proceedings of the IEEE/CVF International Conference on Computer Vision}, 622--631.

\bibitem[{Oliver et~al.(2018)Oliver, Odena, Raffel, Cubuk, and Goodfellow}]{oliver2018realistic}
Oliver, A.; Odena, A.; Raffel, C.~A.; Cubuk, E.~D.; and Goodfellow, I. 2018.
\newblock Realistic evaluation of deep semi-supervised learning algorithms.
\newblock \emph{Advances in neural information processing systems}, 31.

\bibitem[{Oord, Li, and Vinyals(2018)}]{oord2018representation}
Oord, A. v.~d.; Li, Y.; and Vinyals, O. 2018.
\newblock Representation learning with contrastive predictive coding.
\newblock \emph{arXiv preprint arXiv:1807.03748}.

\bibitem[{Papyan, Han, and Donoho(2020)}]{papyan2020prevalence}
Papyan, V.; Han, X.; and Donoho, D.~L. 2020.
\newblock Prevalence of neural collapse during the terminal phase of deep learning training.
\newblock \emph{Proceedings of the National Academy of Sciences}, 117(40): 24652--24663.

\bibitem[{Perronnin, S{\'a}nchez, and Mensink(2010)}]{perronnin2010improving}
Perronnin, F.; S{\'a}nchez, J.; and Mensink, T. 2010.
\newblock Improving the fisher kernel for large-scale image classification.
\newblock In \emph{Computer Vision--ECCV 2010: 11th European Conference on Computer Vision, Heraklion, Crete, Greece, September 5-11, 2010, Proceedings, Part IV 11}, 143--156. Springer.

\bibitem[{Radford et~al.(2021)Radford, Kim, Hallacy, Ramesh, Goh, Agarwal, Sastry, Askell, Mishkin, Clark et~al.}]{radford2021learning}
Radford, A.; Kim, J.~W.; Hallacy, C.; Ramesh, A.; Goh, G.; Agarwal, S.; Sastry, G.; Askell, A.; Mishkin, P.; Clark, J.; et~al. 2021.
\newblock Learning transferable visual models from natural language supervision.
\newblock In \emph{International conference on machine learning}, 8748--8763. PMLR.

\bibitem[{Robbins and Monro(1951)}]{robbins1951stochastic}
Robbins, H.; and Monro, S. 1951.
\newblock A stochastic approximation method.
\newblock \emph{The annals of mathematical statistics}, 400--407.

\bibitem[{Schroff, Kalenichenko, and Philbin(2015)}]{schroff2015facenet}
Schroff, F.; Kalenichenko, D.; and Philbin, J. 2015.
\newblock Facenet: A unified embedding for face recognition and clustering.
\newblock In \emph{Proceedings of the IEEE conference on computer vision and pattern recognition}, 815--823.

\bibitem[{Sohn(2016)}]{sohn2016improved}
Sohn, K. 2016.
\newblock Improved deep metric learning with multi-class n-pair loss objective.
\newblock \emph{Advances in neural information processing systems}, 29.

\bibitem[{Wah et~al.(2011)Wah, Branson, Welinder, Perona, and Belongie}]{wah2011caltech}
Wah, C.; Branson, S.; Welinder, P.; Perona, P.; and Belongie, S. 2011.
\newblock The caltech-ucsd birds-200-2011 dataset.

\bibitem[{Wang and Liu(2021)}]{wang2021understanding}
Wang, F.; and Liu, H. 2021.
\newblock Understanding the behaviour of contrastive loss.
\newblock In \emph{Proceedings of the IEEE/CVF conference on computer vision and pattern recognition}, 2495--2504.

\bibitem[{Wang et~al.(2020)Wang, Sun, Cheng, Jiang, Deng, Zhao, Liu, Mu, Tan, Wang et~al.}]{wang2020deep}
Wang, J.; Sun, K.; Cheng, T.; Jiang, B.; Deng, C.; Zhao, Y.; Liu, D.; Mu, Y.; Tan, M.; Wang, X.; et~al. 2020.
\newblock Deep high-resolution representation learning for visual recognition.
\newblock \emph{IEEE transactions on pattern analysis and machine intelligence}, 43(10): 3349--3364.

\bibitem[{Wang et~al.(2019)Wang, Liew, Zou, Zhou, and Feng}]{wang2019panet}
Wang, K.; Liew, J.~H.; Zou, Y.; Zhou, D.; and Feng, J. 2019.
\newblock Panet: Few-shot image semantic segmentation with prototype alignment.
\newblock In \emph{proceedings of the IEEE/CVF international conference on computer vision}, 9197--9206.

\bibitem[{Wang and Isola(2020)}]{wang2020understanding}
Wang, T.; and Isola, P. 2020.
\newblock Understanding contrastive representation learning through alignment and uniformity on the hypersphere.
\newblock In \emph{International Conference on Machine Learning}, 9929--9939. PMLR.

\bibitem[{Wang et~al.(2021)Wang, Zhou, Yu, Dai, Konukoglu, and Van~Gool}]{wang2021exploring}
Wang, W.; Zhou, T.; Yu, F.; Dai, J.; Konukoglu, E.; and Van~Gool, L. 2021.
\newblock Exploring cross-image pixel contrast for semantic segmentation.
\newblock In \emph{Proceedings of the IEEE/CVF International Conference on Computer Vision}, 7303--7313.

\bibitem[{Wu et~al.(2023)Wu, Guo, Li, Yu, Gao, and Sang}]{wu2023semantic}
Wu, D.; Guo, Z.; Li, A.; Yu, C.; Gao, C.; and Sang, N. 2023.
\newblock Semantic Segmentation via Pixel-to-Center Similarity Calculation.
\newblock \emph{arXiv preprint arXiv:2301.04870}.

\bibitem[{Wu et~al.(2018)Wu, Xiong, Yu, and Lin}]{wu2018unsupervised}
Wu, Z.; Xiong, Y.; Yu, S.~X.; and Lin, D. 2018.
\newblock Unsupervised feature learning via non-parametric instance discrimination.
\newblock In \emph{Proceedings of the IEEE conference on computer vision and pattern recognition}, 3733--3742.

\bibitem[{Xie et~al.(2021)Xie, Wang, Yu, Anandkumar, Alvarez, and Luo}]{xie2021segformer}
Xie, E.; Wang, W.; Yu, Z.; Anandkumar, A.; Alvarez, J.~M.; and Luo, P. 2021.
\newblock SegFormer: Simple and efficient design for semantic segmentation with transformers.
\newblock \emph{NeurIPS}, 34.

\bibitem[{Xu et~al.(2020)Xu, Xian, Wang, Schiele, and Akata}]{xu2020attribute}
Xu, W.; Xian, Y.; Wang, J.; Schiele, B.; and Akata, Z. 2020.
\newblock Attribute prototype network for zero-shot learning.
\newblock \emph{Advances in Neural Information Processing Systems}, 33: 21969--21980.

\bibitem[{Yang et~al.(2022)Yang, Wu, Zhang, Jiang, Liu, Zheng, Zhang, Wang, and Zeng}]{yang2022class}
Yang, F.; Wu, K.; Zhang, S.; Jiang, G.; Liu, Y.; Zheng, F.; Zhang, W.; Wang, C.; and Zeng, L. 2022.
\newblock Class-aware contrastive semi-supervised learning.
\newblock In \emph{Proceedings of the IEEE/CVF Conference on Computer Vision and Pattern Recognition}, 14421--14430.

\bibitem[{Yuan, Chen, and Wang(2020)}]{yuan2020object}
Yuan, Y.; Chen, X.; and Wang, J. 2020.
\newblock Object-contextual representations for semantic segmentation.
\newblock In \emph{Computer Vision--ECCV 2020: 16th European Conference, Glasgow, UK, August 23--28, 2020, Proceedings, Part VI 16}, 173--190. Springer.

\bibitem[{Zhang et~al.(2021)Zhang, Li, Yan, He, and Sun}]{zhang2021distribution}
Zhang, S.; Li, Z.; Yan, S.; He, X.; and Sun, J. 2021.
\newblock Distribution alignment: A unified framework for long-tail visual recognition.
\newblock In \emph{Proceedings of the IEEE/CVF conference on computer vision and pattern recognition}, 2361--2370.

\bibitem[{Zhong et~al.(2021)Zhong, Cui, Liu, and Jia}]{zhong2021improving}
Zhong, Z.; Cui, J.; Liu, S.; and Jia, J. 2021.
\newblock Improving calibration for long-tailed recognition.
\newblock In \emph{Proceedings of the IEEE/CVF conference on computer vision and pattern recognition}, 16489--16498.

\bibitem[{Zhou et~al.(2017)Zhou, Zhao, Puig, Fidler, Barriuso, and Torralba}]{zhou2017scene}
Zhou, B.; Zhao, H.; Puig, X.; Fidler, S.; Barriuso, A.; and Torralba, A. 2017.
\newblock Scene parsing through ade20k dataset.
\newblock In \emph{Proceedings of the IEEE conference on computer vision and pattern recognition}, 633--641.

\bibitem[{Zhou et~al.(2022)Zhou, Wang, Konukoglu, and Van~Gool}]{zhou2022rethinking}
Zhou, T.; Wang, W.; Konukoglu, E.; and Van~Gool, L. 2022.
\newblock Rethinking semantic segmentation: A prototype view.
\newblock In \emph{Proceedings of the IEEE/CVF Conference on Computer Vision and Pattern Recognition}, 2582--2593.

\end{thebibliography}
